  \documentclass{gretsi}

   \usepackage[english,french]{babel}   % "babel.sty" + "french.sty"
  \usepackage{times}			% ajout times le 30 mai 2003
 
\usepackage{epsfig}
\usepackage{graphicx}
\usepackage{amsmath}
\usepackage{amssymb}
 \usepackage{booktabs}
 \usepackage{multirow}
\usepackage{pifont}
\usepackage{caption}
\usepackage{subcaption}

\usepackage{array}
\usepackage{color}
\usepackage{colortbl}

\usepackage{pifont}
\usepackage{amssymb}
\usepackage{latexsym}

\title{Comparing feature fusion strategies for \\ Deep Learning-based kidney stone identification}

\author{\coord{Elias}{Villalvazo-Avila}{1},
       \coord{Francisco}{Lopez-Tiro}{1},
        \coord{Daniel}{Flores-Araiza}{1},
        \coord{Gilberto}{Ochoa-Ruiz}{1},\\
        \coord{Jonathan}{El-Beze}{2},
        \coord{Jacques}{Hubert}{2},
        \coord{Christian}{Daul}{3}}

\address{\affil{1}{Escuela de Ingenieria y Ciencias, Tecnologico de Monterrey \\
         Av. Eugenio Garza Sada 2501 Sur, Tecnológico, 64849 Monterrey, N.L., Mexico}
         \affil{2}{CHU Nancy, Service d’urologie de Brabois, F-54511 Nancy, France}
         \affil{3}{CRAN (UMR 7039), Université de Lorraine and CNRS, \\ 2 avenue de la For\^et de Haye, 54518 Vandœuvre-l\`es-Nancy cedex, France}}

\email{A01798097, A01799045, A01098051, gilberto.ochoa@tec.mx \\  j.elbeze@chru-nancy.fr,
j.hubert@chru-nancy.fr, christian.daul@univ-lorraine.fr}

\frenchabstract{Cette contribution présente une méthode d’apprentissage profond pour l’extraction et la fusion d’informations d’images acquises sous différents points de vue dans le but de produire des caractéristiques plus discriminantes entre objets. Notre approche a été conçue pour mimer l'analyse morpho-constitutionnelle utilisée par les urologues pour classer visuellement des fragments de calculs rénaux à partir de leur surface et section. Des stratégies de fusion de caractéristiques profondes ont permis d’améliorer les performances des extracteurs (structures principales des réseaux) à vue unique de plus de 10 \% en termes de précision de la classification des calculs rénaux.}

\englishabstract{This contribution presents a deep-learning method for extracting and fusing image information acquired from different viewpoints with the aim to produce more discriminant object features. Our approach was specifically designed to mimic the morpho-constitutional analysis used by urologists to visually classify kidney stones by inspecting the sections and surfaces of their fragments. Deep feature fusion strategies improved the results of single view extraction backbone models by more than 10\% in terms of precision of the kidney stones classification.}

\begin{document}
\maketitle
\section{Introduction}
\vspace*{-3mm}
Urolithiasis refers to the formation of kidney stones that cannot be expelled from the urinary tract. 
This is a medical condition that has been increasing over the last few years \cite{viljoen2019renal}. 
Urolithiasis is caused by multiple factors, where diet is the most important, but also genetic inheritance, water intake, and a sedentary lifestyle could promote the formation of kidney stones \cite{friedlander2015diet}.
The Morpho-constitutional analysis (MCA) is the most important method for kidney stone characterisation. MCA is a combination of a visual examination under the microscope of the stone’s texture, appearance, and color (surface and section views), and a biochemical analysis by Fourier Transform Infrared Spectroscopy (FTIR) \cite{daudon2018recurrence}. If carried out properly, a timely treatment (diet adaptation, surgery) can be prescribed for each patient, reducing the risk of stone recurrence \cite{daudon2018recurrence}.

\begin{table*}[]
\centering
\caption{Overview of the kidney stone classes and acquisition conditions in the-state-of-the-art works, as well as for this contribution.  Simplified taxonomy: UA: Uric Acid (Anhydrous and Dihydrate), WW: Whewellite (Calcium Oxalate Monohydrate), WD: Weddellite (Calcium Oxalate Dihydrate), STR: Struvite, CYS: Cystine, and BRU: Brushite.}
\arrayrulecolor{black}
\scalebox{0.85}{%
\begin{tabular}{!{\color{black}\vrule}l!{\color{black}\vrule}l!{\color{black}\vrule}l!{\color{black}\vrule}l!{\color{black}\vrule}l!{\color{black}\vrule}l!{\color{black}\vrule}l!{\color{black}\vrule}l!{\color{black}\vrule}l!{\color{black}\vrule}l!{\color{black}\vrule}} 
\hline
\multirow{2}{*}{Reference/Feature} & \multicolumn{6}{l!{\color{black}\vrule}}{Kidney Stone Composition} & \multicolumn{2}{l!{\color{black}\vrule}}{Image Type} & \multirow{2}{*}{Acquisition}  \\ 
\cline{2-9}
                                   & AU         & WW         & WD          & STR         & CYS           & BRU           & Surface        & Section  &          \\ 
\hline
{}Serrat et al. (2017) \cite{serrat2017mystone}              & \ding{51}  & \ding{51}   & \ding{51}    & \ding{51}   &  &                  & \ding{51}    &  \ding{51}        &      Ex-vivo     \\ 
\cline{1-1}
{}Torrel et al. (2018) \cite{torrell2018metric}             & \ding{51}  & \ding{51}   & \ding{51}    & \ding{51}   &     & \ding{51}   & \ding{51}      &           &    Ex-vivo        \\ 
\cline{1-1} 
{}Black et al. (2020) \cite{black2020deep}              & \ding{51}  & \ding{51}   &              & \ding{51}   & \ding{51}      & \ding{51}    & \ding{51}      & \ding{51}  &   Ex-vivo          \\ 
\cline{1-1}
{}Martinez et al. (2020) \cite{martinez2020towards}               &         \ding{51}  & \ding{51}   &  \ding{51}       &   &     &    &  \ding{51}    & \ding{51}  &   In-vivo          \\ 
\cline{1-1}
{}Lopez et al. (2021) \cite{lopez2021assessing}               &         \ding{51}  & \ding{51}   &  \ding{51}       &   &     & \ding{51}    &  \ding{51}    & \ding{51}  &   In-vivo          \\ 
\cline{1-1}
{}Estrade et al. (2021) \cite{estrade2021}               &         \ding{51}  & \ding{51}   &  \ding{51}       &   &     &    &  \ding{51}    & \ding{51}  &   In-vivo          \\ 
\cline{1-1}
This contribution                               & \ding{51}  & \ding{51}   & \ding{51}     &      \ding{51}      &         \ding{51}      & \ding{51}    & \ding{51}       &         \ding{51}        &           Ex-vivo           \\
\hline
\end{tabular}
}
\arrayrulecolor{black}
\label{pworks1}
\end{table*}

\begin{table*}[]
\centering
\caption{Comparison of the  precision obtained by the three main contributions for the most common kidney stone classes. The precision is given for each individual class and classifier. The taxonomy per stone class same as in Table \ref{pworks1}. The average precision (weighted by the image number of each class) are also given for each kidney stone type. In the last table line, the precision values are given for surface images. For section images, the authors in \cite{estrade2021}, reported a precision of 0.95, 0.94 and 0.94 for AU, WW and WD, respectively. The corresponding weighted precision equals 0.94.}
\scalebox{0.85}{%
\begin{tabular}{|l|c|c|c|c|c|c|c|c|}
\hline
\multirow{2}{*}{Reference} & \multicolumn{6}{c|}{Precision Per Class} & \multirow{2}{*}{\begin{tabular}[c]{@{}c@{}}Weighted \\ Precision\end{tabular}} & \multirow{2}{*}{ML Method} \\ \cline{2-7}
 & AU & WW & WD & STR & CYS & BRU &  &  \\ \hline
{}Serrat et al. (2017) \cite{serrat2017mystone}{} & 0.65 & 0.55 & 0.69 & 0.50 & N/A & N/A & 0.63 & Random Forest \\ \hline
{}Torrel et al. (2018) \cite{torrell2018metric}{} & 0.76 & 0.67 & 0.80 & 071 & N/A & 0.72 & 0.74 & Siamese CNN \\ \hline
{}Black et al. (2020) \cite{black2020deep}{} & 0.94 & 0.95 & N/A & 0.71 & 0.75 & 0.75 & 0.85 & CNN – ResNet101 \\ \hline
{}Martinez et al. (2020) \cite{martinez2020towards}& 0.91 & 0.94 & 0.92 & N/A & N/A & N/A & 0.92 & Random Forest \\ \hline
{}Lopez et al. (2021) \cite{lopez2021assessing} & 0.98 & 0.93 & 0.95 & N/A & N/A & 0.96 & 0.95 & Inception \\ \hline
{}Estrade et al. (2021) \cite{estrade2021} & 0.99 & 0.90 & 0.93 & N/A & N/A & N/A & 0.94 & ResNet152v2 \\ \hline
\end{tabular}
}
\label{pworks2}
\end{table*}

However, MCA has a major drawback: the results of this analysis are often available only after several weeks. Thus, urologists increasingly aim at visually identifying the morphology of kidney stones only with the help of the image displayed on the screen \cite{ochoa2022vivo} during the removal process (Endoscopic Stone recognition (ESR)). However, this visual analysis requires a great deal of experience due to the high similarities between classes that only a limited number of specialists have. %\cite{bergot2019base}. 

Therefore, different Machine Learning (ML) approaches have been proposed \cite{ochoa2022vivo, estrade2021, martinez2020towards, serrat2017mystone} for the classification of kidney stones, demonstrating that it is a problem that can be solved with traditional and deep learning techniques with very encouraging results.
However, most of these models were trained on ex-vivo stones placed in controlled environments, whereas in reality, images may suffer from motion blur, reflections, illumination variations, as occurs in common practice during an endoscopic imaging session. Moreover, there is no ordered manner of mixing surface and section information for exploiting the visual information in a way that a specialist would do it. Besides, in most cases the amount of training data available is limited, thus these contributions use data augmentation techniques to increase the amount of input data, but some limitations have not been addressed. Nonetheless, these works \cite{ochoa2022vivo, estrade2021, martinez2020towards} have demonstrated the potential of automatic ESR in an in-vivo dataset. Table \ref{pworks1} gathers the previous works which have addressed the problem of recognizing kidney stone types using only images. This table provides also an overview on the used data (urinary calculus classes and acquisition conditions).

Multi-View (MV) classification is an area of ML that combines features from different sources or feature subsets, known as views, to identify objects with higher accuracy, since diverse characteristics are extracted,  synthesized and combined \cite{2016multiview}. 

This variant of learning can improve the performance by optimizing multiple functions, one per view, and in that way, information can be obtained from different perspectives of the same data inputs. Moreover, MV can also be applied to Convolutional Neural Networks (CNNs) to boost the performance in situations where a single image does not yield sufficiently discriminative information for accurate classification by combining useful information from different views, so more comprehensive representations may be learned yielding a more effective classifier. %\cite{seeland_2021}.
% Single view 
This MV approach stands in contrast to previous works for ML-based ESR, which made use of single models which mixed image patches to train either a shallow or a DL models, as summarized in Table \ref{pworks2}. These previous works showed the potential gains that could be obtained through an automated mechanisms for ESR, but those methods either used ex-vivo images acquired using high-quality microscopes or digital cameras or did not take into account the fusion of section and surface information to perform a prediction.

In this work, we leverage recent strides in DL research that have sought to combine information from multiple views and  we demonstrate that MV learning can be applied to  ESR, with promising results. %\cite{geras2018highresolution, 2016pulmonary_multiview} 
We show that such an approach can be beneficial for the identification of kidney stones by maximizing the amount of information that the model can use for classification. Through several experiments, we demonstrate that by combining the information of surface and section views in the same model, we can obtain a method that is more explainable and similar to what specialists do in clinical practice (i.e., MCA).

\section{Materials and Methods}

\subsection{Kidney stone dataset}

\begin{figure} [t!]
     \centering
     \includegraphics[width=0.90\linewidth]{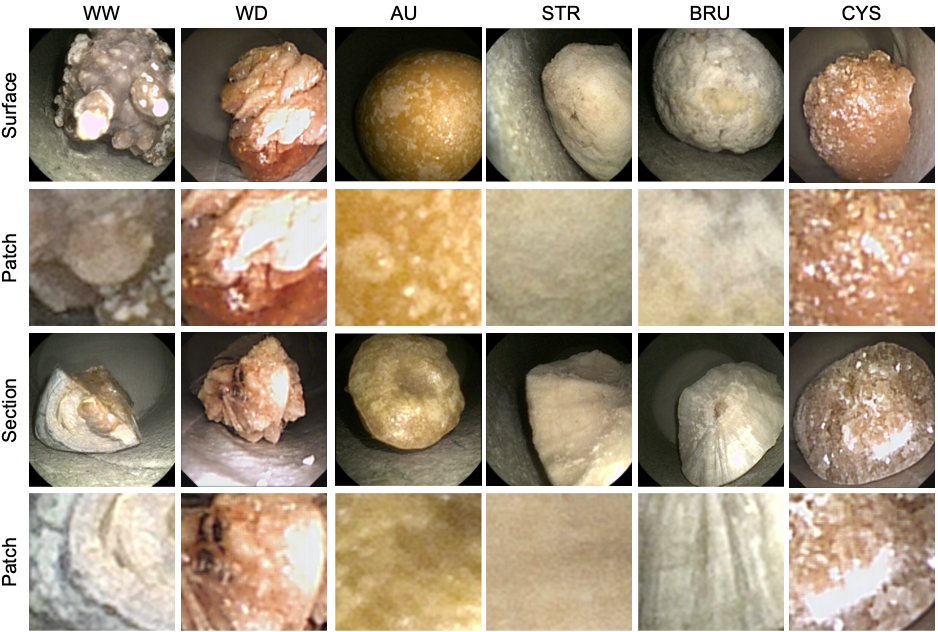}
   \caption{Examples of ex-vivo kidney stones images. From left to right: WW, WD, AU, STR, BRU, and CYS. The surface view is in the first row, and their respective generated patches are in the second row. The section view is in the third row, and their respective generated patches are in the fourth row.}
     \label{fig:im1}
     \end{figure}

The ex-vivo dataset includes 305 kidney stone images acquired (two reusable digital flexible ureteroscopes from Karl Storz using video columns: Storz Image 1 Hub and Storz image1 S) and labeled manually by the urologist Jonathan El Beze$^{2}$. 
To reproduce in-vivo conditions, the experimental setup used in this work consists of a small diameter tube where the inner walls were covered with a yellowish film to display the appearance of the urinary tract (for more details, see \cite{elbeze2022}).
The ex-vivo dataset consists of three subsets: the first subset consist of 177 surface images, 128 section images for the second subset, and the third subset of 305 images (177 section + 128 surface) of the six kidney stone types with the highest incidence: Type Ia (Whewellite, WW), Type IIb (Weddellite, WD), Type IIIb (Acide Urique, AU), Type IVc (Struvite, STR), Type IVd (Brushite, BRU), and Type Va (Cystine, CYS). Images of this dataset are shown in Fig. \ref{fig:im1}.

Classification of kidney stones (i.e., MCA) is usually  not performed on whole images \cite{martinez2020towards, black2020deep, torrell2018metric, serrat2017mystone}. Therefore, in this work as in previous works, patches of 256$\times$256 pixels were cropped from the original images to increase the size of the training dataset (for more details, see \cite{lopez2021assessing}). 
However, the number of resultant patches for each class is imbalanced (due to the changing fragment sizes, image resolution, and the number of images in the original dataset). In order to balance the number of patches per class, a random sampling approach was used. 

This step yielded a total of 1000 patches per class (WW, WD, AU, STR, BRU, and CYS) and view (surface, section, and mixed). This new dataset was then split into 19200 images ($80\%$) for training and validation, and 4800 images ($20\%$) for the test.
In order to limit the over-fitting produced by the small size of the available training dataset, data augmentation was heavily performed. Additional patches were obtained by applying geometrical transformations (patch flipping, affine transformations, and perspective distortions). The number of patches increased from 19200 to 153600 using data augmentation ($10\%$ of the original patches were kept for test purposes). The patches were also ``whitened" using the mean $m_{i}$ and standard deviation $\sigma_{i}$ of the color values $I_{i}$ in each channel \cite{lopez2021assessing}.

%$(I^{w}_{i} = (I_{i}-m_{i}\sigma_{i})$, with $i=R,G,B)$. 

\subsection{Methods}

\subsubsection{Pre-training Stage}
Previous approaches have used Deep Learning architectures such as AlexNet, or VGG16 for assessing the kidney stone classification task \cite{ochoa2022vivo, lopez2021assessing}. For this contribution, the previously-mentioned architectures are used for the creation of the MV models. This network was trained on the entire training data, mixing both surface and section patches, and served as a baseline or comparison for the multi-view implementations introduced in this paper. Once trained, the feature extraction layers of this single-view network are frozen to ensure that each branch from the multi-view model extracts the same features and that any variation in the performance will rely on the unfrozen layers (fusion and fully-connected layers).

 \begin{figure} [h!]
     \centering
     \includegraphics[width=0.65\linewidth]{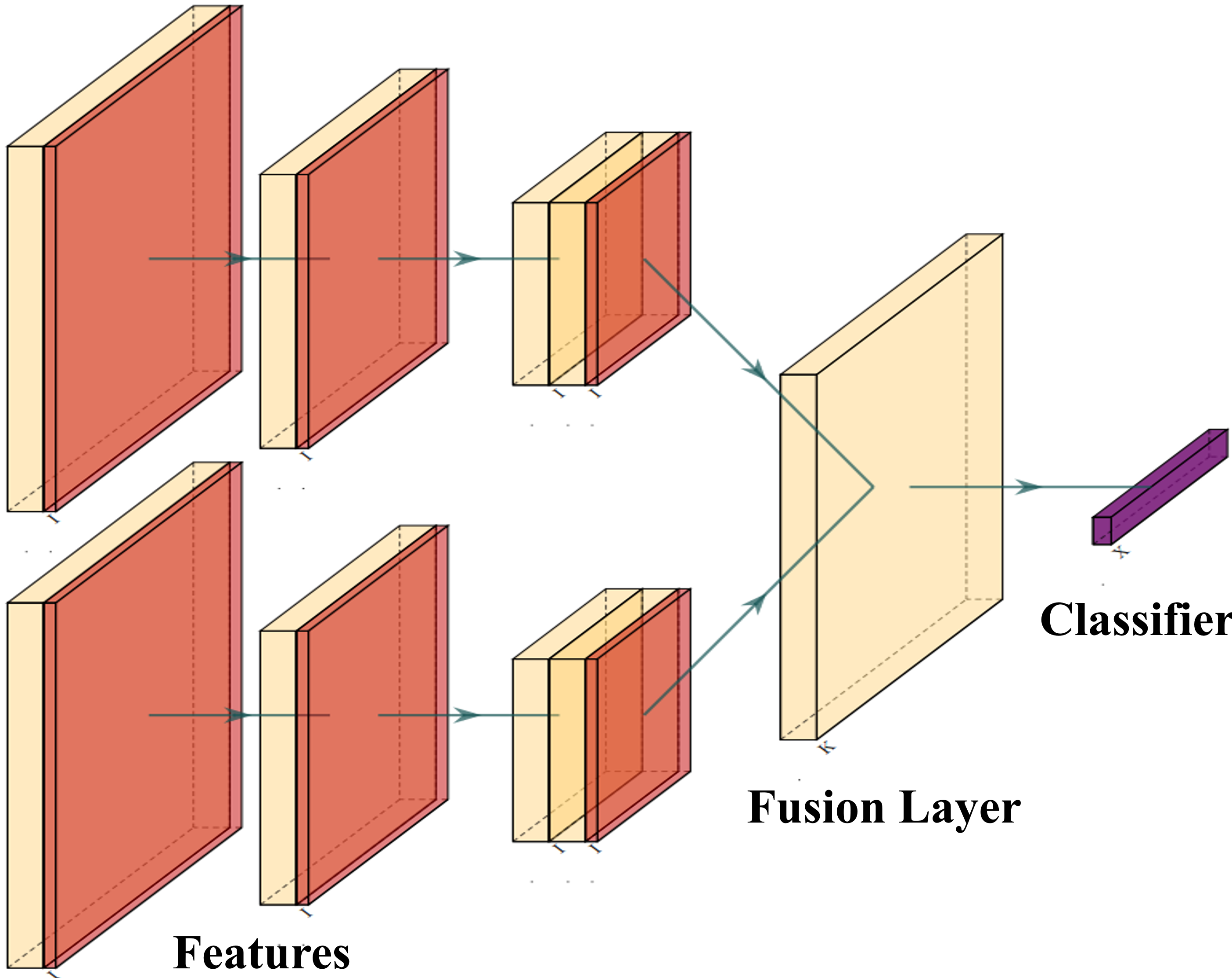}
   \caption{Proposed Multi-View model. First part of the model corresponds to the duplicated feature extraction layers. One copy will process only surface images, the other will process section images. A fusion layer is added to combine information from the different views. The fused feature map is then fed to the classification layers }
     \label{fig:multi_model}
     \end{figure}

\subsubsection{Multi-View Model}
The frozen layers are duplicated. In this way, one copy will process only images of the surface of the stone, while the other copy will process images of the section view. These frozen layers are connected to a fusion layer, which will be responsible for mixing the information of the two views. In this work, the two late-fusion methods proposed in \cite{2016multiview} are explored. The first method concatenates the feature vectors obtained from each view, and connects the resulting representation to a fully connected layer. As per the second method, feature vectors are stacked and max-pooling is applied across them.  Lastly, the output of the fusion layer is connected to the remaining part of the MV model, which merely consists of the classifier. The proposed model is shown in Fig. \ref{fig:multi_model}.
To make a direct comparison of the feasibility and performance of this architecture against previous works, we used the same hyper-parameters as \cite{ochoa2022vivo}. Cross-entropy loss is used to compute the classification loss, and optimization is performed using Adam optimizer with a learning rate of $2e^{-4}$. Batch-size selected was 64 for both multi-view and single-view networks. 

%The experiments and implementation of this network were performed using Pytorch v1.10.2.

 \begin{figure*} [h!] 
     \centering
     \subfloat[SV-AlexNet]{
     \label{fig:umap_alex_sv}
     \includegraphics[width=0.34\linewidth]{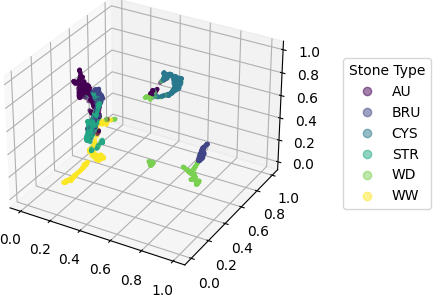}}
     \hspace{15mm}
     \subfloat[MV-AlexNet-max]{
     \label{fig:umap_alex_mvpool}
     \includegraphics[width=0.34\linewidth]{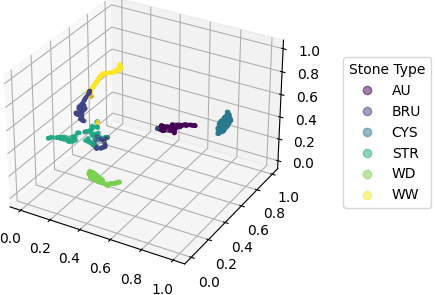}}
     \caption{UMAP visualizations showing the learned representations (i.e., feature maps) for  (a) the Single-View (SV) AlexNet architecture with no fusion, and (b) in Multi-View AlexNet architecture with a max-pool fusion layer. As it can be observed in the figure, more separate clusters are obtained through the use of feature fusion layers }
 \label{fig:umaps}
     \end{figure*}
     
\section{Results and Discussion}
Several experiments were performed to assess the ability of MV models to predict the kidney stone class, combining information from surface and section views, as done during an ESR procedure \cite{estrade2021}. Precision (P) and Recall (R) metrics are determined for each class individually. The results reported for both fusion techniques in the multi-view models show that combining information from different classifiers yields significant results compared to the single-view classifiers. For instance, for a single-view AlexNet network, the results obtained were 0.84 and 0.83 for precision and recall, respectively. In contrast, MV networks, independent of the fusion strategy, performed better compared to the other experiments. Table \ref{tab:Results} shows the scores for all the models used for this work, and the UMAP visualizations on Fig. \ref{fig:umap_alex_sv}, and Fig. 
\ref{fig:umap_alex_mvpool} show how stone type clusters are distributed for both SV and MV networks. One disadvantage of using concatenation as fusion strategy is that the number of features of the first layers of the classifier increases considerably, limiting its implementation on systems with reduced memory.

 \begin{table}[]
 \normalsize
 \centering
 \caption{Weighted average metrics comparison for section, surface, and mixed patches. MV-AlexNet-max: Multi-View network with max-pool as fusion strategy. MV-VGG16-max: Multi-View network with max-pool as fusion strategy. MV-AlexNet-conc: Multi-View network with concatenation as fusion strategy.  SV-AlexNet: Single-View AlexNet network. SV-VGG16: Single-View VGG16 network.}
 \label{tab:Results}
 \scalebox{0.85}{%
 \begin{tabular}{@{}ccccccc@{}}
 \cmidrule(l){2-7}
                 & \multicolumn{2}{c}{Surface} & \multicolumn{2}{c}{Section} & \multicolumn{2}{c}{Mixed}     \\ \midrule
 Classifier      & \textbf{P}   & \textbf{R}   & \textbf{P}   & \textbf{R}   & \textbf{P}    & \textbf{R}    \\ \midrule
 MV-AlexNet-max  & --            & --            & --            & --            & \textbf{0.95} & \textbf{0.94} \\
 MV-VGG16-max    & --            & --            & --            & --            & \textbf{0.94} & \textbf{0.94} \\
 MV-AlexNet-conv & --            & --            & --            & --            & \textbf{0.94} & \textbf{0.93} \\
 SV-AlexNet      & 0.77         & 0.71         & 0.88         & 0.87         & 0.84          & 0.83          \\
 SV-VGG16        & 0.79         & 0.70         & 0.89            & 0.89            & 0.83          & 0.81          \\
 \bottomrule
 \end{tabular}%
 }
 \end{table}

\section{Conclusion and future work}
\label{discussion_future_work}
We showed that by mixing information from different views, it is possible to train more accurate models for predicting kidney stone composition from images obtained from ureteroscopy. Thus, AI technology can be included in the current stone removal workflow, speeding up preventive diagnosis measures. However, we make use of a very-limited ex-vivo dataset in a simulated environment. We aim to solve this problem by applying metric learning in future work to tackle the amount of data that we require for training, as well as to increase inter-class separability.

\end{document}